\documentclass[10pt,twocolumn,letterpaper]{article}

\usepackage[pagenumbers]{cvpr} % To force page numbers, e.g. for an arXiv version

\usepackage{graphicx}
\usepackage{amsmath}
\usepackage{amssymb}
\usepackage{booktabs}
\usepackage{stfloats}
\usepackage{cite}
\usepackage{color}
\usepackage{times}
\usepackage{overpic}
\usepackage{bm}
\usepackage{tabu}
\usepackage{bbding}
\usepackage{multicol}

\usepackage{diagbox}
\usepackage{multirow}
\usepackage{scalerel,stackengine}
\usepackage[table ]{xcolor}
\usepackage[accsupp]{axessibility}
\newcommand{\PAR}[1]{\vskip3pt \noindent {{\bf #1~}}}
\newcommand{\TITLE}{SLOPER4D}
\newcommand{\framevideo}{300k~}
\newcommand{\framelidar}{100k~}
\newcommand{\framemocap}{500k~}
\newcommand{\numberscene}{10~}
\newcommand{\numberseq}{15~}
\newcommand{\numberperson}{12~}

\usepackage[pagebackref,breaklinks,colorlinks]{hyperref}
\usepackage[capitalize]{cleveref}
\crefname{section}{Sec.}{Secs.}
\Crefname{section}{Section}{Sections}
\Crefname{table}{Table}{Tables}
\crefname{table}{Tab.}{Tabs.}

\def\cvprPaperID{5047} % *** Enter the CVPR Paper ID here
\def\confName{CVPR}
\def\confYear{2023}

\begin{document}

\title{SLOPER4D: A Scene-Aware Dataset for Global 4D Human Pose Estimation in Urban Environments

}

\author{Yudi~Dai\textsuperscript{1}
\and Yitai~Lin\textsuperscript{1}
\and Xiping~Lin\textsuperscript{1}
\and Chenglu~Wen\textsuperscript{1}\thanks{Corresponding author.}
\and Lan~Xu\textsuperscript{2}
\and Hongwei~Yi\textsuperscript{3} 
\and Siqi~Shen\textsuperscript{1} 
\and Yuexin~Ma\textsuperscript{2}
\and Cheng~Wang\textsuperscript{1}
\and $^{1}$Xiamen University, China 
\hspace{20mm}$^{2}$ShanghaiTech University, China
\and
$^{3}$Max Planck Institute for Intelligent Systems, Germany
}

\makeatletter
\let\@oldmaketitle\@maketitle% Store \@maketitle
\renewcommand{\@maketitle}{
   \@oldmaketitle% Update \@maketitle to insert...
	\begin{center}
      \vspace{-6mm}
      \includegraphics[width=0.96\linewidth]{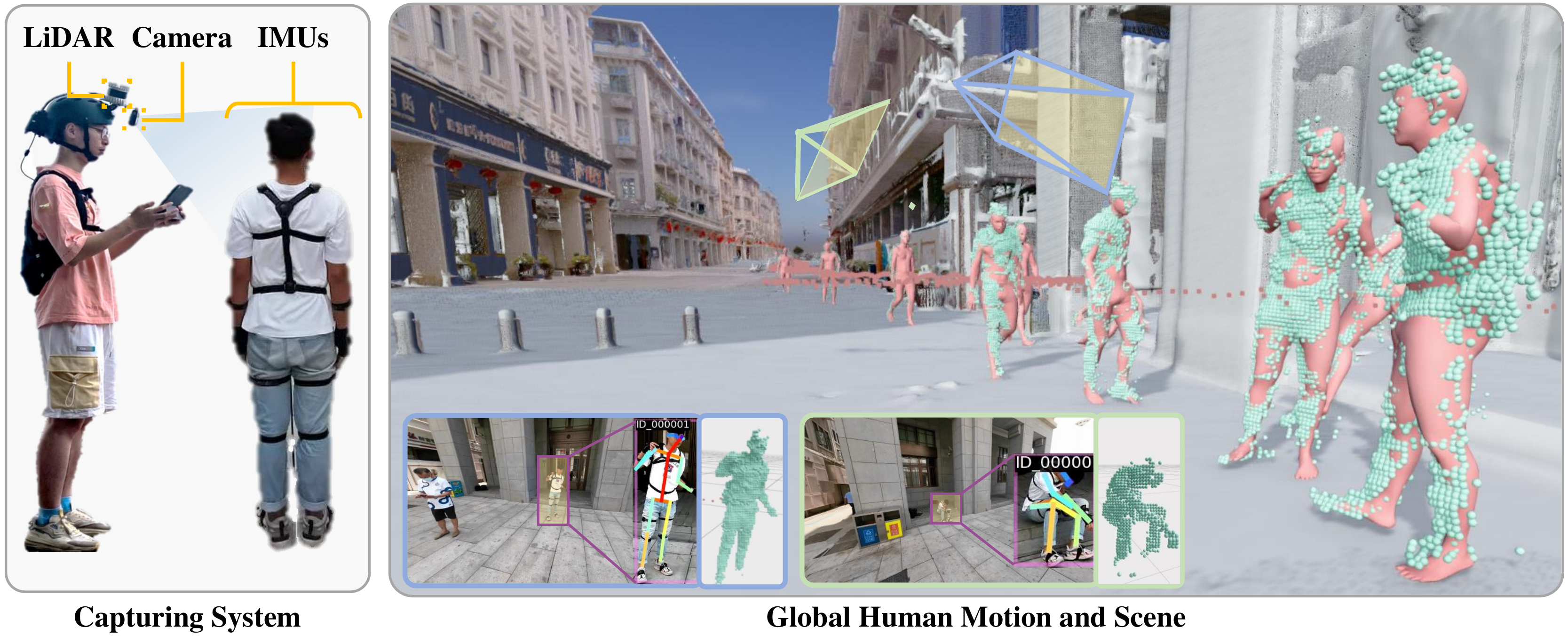}
	\end{center}
   \vspace{-2mm}

  \refstepcounter{figure}\normalfont Figure~\thefigure. 
  Using the head-mounted LiDAR and camera to scan the IMUs wearer, we construct \TITLE, a large scene-aware dataset for global 4D human pose estimation in urban environments, including LiDAR point clouds, the RGB videos with 2D/3D annotations, accurate global human pose annotations, and the reconstructed scene.
  \label{fig:teaser}
  \newline
  }

\makeatother

\maketitle

\begin{abstract}

   We present SLOPER4D, a novel scene-aware dataset collected in large urban environments to facilitate the research of global human pose estimation (GHPE) with human-scene interaction in the wild. Employing a head-mounted device integrated with a LiDAR and camera, we record 12 human subjects' activities over 10 diverse urban scenes from an egocentric view. Frame-wise annotations for 2D key points, 3D pose parameters, and global translations are provided, together with reconstructed scene point clouds. To obtain accurate 3D ground truth in such large dynamic scenes, we propose a joint optimization method to fit local SMPL meshes to the scene and fine-tune the camera calibration during dynamic motions frame by frame, resulting in plausible and scene-natural 3D human poses. Eventually, SLOPER4D consists of 15 sequences of human motions, each of which has a trajectory length of more than 200 meters (up to 1,300 meters) and covers an area of more than 2,000 $m^2$ (up to 13,000 $m^2$), including more than 100K LiDAR frames, 300k video frames, and 500K IMU-based motion frames. With SLOPER4D, we provide a detailed and thorough analysis of two critical tasks, including camera-based 3D HPE and LiDAR-based 3D HPE in urban environments, and benchmark a new task, GHPE. The in-depth analysis demonstrates SLOPER4D poses significant challenges to existing methods and produces great research opportunities. The dataset and code are released at \url{http://www.lidarhumanmotion.net/sloper4d/}.
\end{abstract}

\vspace{-2ex}

\section{Introduction}
\label{sec:intro}

Urban-level human motion capture is attracting more and more attention, which targets at acquiring consecutive fine-grained human pose representations, such as 3D skeletons and parametric mesh models, with accurate global locations in the physical world. It is essential for human action recognition, social-behavioral analysis, and scene perception and further benefits many downstream applications, including Augmented/Virtual Reality, simulation, autonomous driving, smart city, sociology, etc. However, capturing extra large-scale dynamic scenes and annotating detailed 3D representations for humans with diverse poses is not trivial.

Over the past decades, a large number of datasets and benchmarks have been proposed and have greatly promoted the research in 3D human pose estimation (HPE). They can be divided into two main categories according to the capture environment. The first class usually leverages marker-based systems~\cite{ionescu2013human3, sigal2010humaneva,mahmood2019amass}, cameras~\cite{yi2022human, huang2022capturing,yi2022mime}, or RGB-D sensors~\cite{hassan2019prox, SiweiZhang2021EgoBodyHB} to capture human local poses in constrained environments. However, the optical system is sensitive to light and lacks depth information, making it unstable in outdoor scenes and difficult to provide global translations, and the RGB-D sensor has limited range and could not work outdoors. The second class~\cite{Marcard_2018_ECCV, roetenberg2009xsens} attempts to take advantage of body-mounted IMUs to capture occlusion-free 3D poses in free environments. However, IMUs suffer from severe drift for long-term capturing, resulting in misalignments with the human body. Then, some methods exploit additional sensors, such as RGB camera~\cite{kaichi2020resolving}, RGB-D camera~\cite{xu2017flycap, TrumbleBMVC17,zherong2018}, or LiDAR~\cite{li2022lidarcap} to alleviate the problem and make obvious improvement. However, they all focus on HPE without considering the scene constraints, which are limited for reconstructing human-scene integrated digital urban and human-scene natural interactions.

To capture human pose and related static scenes simultaneously, some studies use wearable IMUs and body-mounted camera~\cite{guzov2021human} or LiDAR~\cite{Dai_2022_CVPR} to register the human in large real scenarios and they are promising for capturing human-involved real-world scenes. However, human pose and scene are decoupled in these works due to the ego view, where auxiliary visual sensors are used for collecting the scene data while IMUs are utilized for obtaining the 3D pose. Different from them, we propose a novel setting for human-scene capture with wearable IMUs and global-view LiDAR and camera, which can provide multi-modal data for more accurate 3D HPE.

In this paper, we propose a huge scene-aware dataset for sequential human pose estimation in urban environments, named SLOPER4D. To our knowledge, it is the first urban-level 3D HPE dataset with multi-modal capture data, including calibrated and synchronized IMU measurements, LiDAR point clouds, and images for each subject. Moreover, the dataset provides rich annotations, including 3D poses, SMPL~\cite{smpl2015loper} models and locations in the world coordinate system, 2D poses and bounding boxes in the image coordinate system, and reconstructed 3D scene mesh. In particular, we propose a joint optimization method for obtaining accurate and natural human motion representations by utilizing multi-sensor complementation and scene constraints, which also benefit global localization and camera calibration in the dynamic acquisition process. Furthermore, SLOPER4D consists of over \numberseq sequences in \numberscene scenes, including library, commercial street, coastal runway, football field, landscape garden, etc., with 2k$\sim$13k $m^2$ area size and $200\sim1,300m$ trajectory length for each sequence. By providing multi-modal capture data and diverse human-scene-related annotations, \TITLE~opens a new door to benchmark urban-level HPE.

{We conduct extensive experiments to show the superiority of our joint optimization approach for acquiring high-quality 3D pose annotations.} Additionally, based on our proposed new dataset, we benchmark two critical tasks: camera-based 3D HPE and LiDAR-based 3D HPE, as well as provide benchmarks for GHPE.

Our contributions are summarized as follows: 
\vspace{-2mm}
\begin{itemize}

    \item We propose the first large-scale urban-level human pose dataset with multi-modal capture data and rich human-scene annotations.

    \item We propose an effective joint optimization method for acquiring accurate human motions in both local and global by integrating LiDAR SLAM results, IMU poses, and scene constraints.

    \item We benchmark two HPE tasks as well as a GHPE task on SLOPER4D, demonstrating its potential of promoting urban-level 3D HPE research.

\end{itemize}

\section{Related Work}
\label{sec:Related work}

\begin{table*}[htb]
     \centering
     \footnotesize

\begin{tabular}{rccccccrrrr}
    \toprule
     Dataset & In the wild & Global & 3D Scene & Point cloud & Video & IMU  & \# Scene  & \# Area size ($m^2$) & \# Subject & \# Frame \\
    \midrule
    \rowcolor[rgb]{ .949,  .949,  .949} H3.6M~\cite{ionescu2013human3} & \textcolor{red}{\XSolidBrush} & \textcolor{green}{\Checkmark} & \textcolor{red}{\XSolidBrush} & \textcolor{red}{\XSolidBrush} & \textcolor{green}{\Checkmark} & \textcolor{red}{\XSolidBrush} & -    & 12   & 11   & 3.6M \\
    3DPW~\cite{Marcard_2018_ECCV} & \textcolor{green}{\Checkmark} & \textcolor{red}{\XSolidBrush} & \textcolor{red}{\XSolidBrush} & \textcolor{red}{\XSolidBrush} & \textcolor{green}{\Checkmark} & \textcolor{green}{\Checkmark} & -    & $<$ 300 & 7    & 51k \\
    \rowcolor[rgb]{ .949,  .949,  .949} PROX~\cite{hassan2019prox} & \textcolor{red}{\XSolidBrush} & \textcolor{green}{\Checkmark} & \textcolor{green}{\Checkmark} & \textcolor{red}{\XSolidBrush} & \textcolor{green}{\Checkmark} & \textcolor{red}{\XSolidBrush} & 12   & $<$ 30 & 20   & 20k \\
    HPS~\cite{guzov2021human} & \textcolor{green}{\Checkmark} & \textcolor{green}{\Checkmark} & \textcolor{green}{\Checkmark} & \textcolor{red}{\XSolidBrush} & \textcolor{green}{\Checkmark}* & \textcolor{green}{\Checkmark} & 8    & 300 $\sim$ 1k & 7    & 7k \\
    \rowcolor[rgb]{ .949,  .949,  .949} HSC4D~\cite{Dai_2022_CVPR} & \textcolor{green}{\Checkmark} & \textcolor{green}{\Checkmark} & \textcolor{green}{\Checkmark} & \textcolor{green}{\Checkmark}* & \textcolor{red}{\XSolidBrush} & \textcolor{green}{\Checkmark} & 5    & 1k $\sim$ 5k & 2    & 10k \\
    LH26M~\cite{li2022lidarcap} & \textcolor{red}{\XSolidBrush} & \textcolor{red}{\XSolidBrush} & \textcolor{red}{\XSolidBrush} & \textcolor{green}{\Checkmark} & \textcolor{green}{\Checkmark} & \textcolor{green}{\Checkmark} & -    & $<$ 200 & 13   & 184k \\
    \rowcolor[rgb]{ .949,  .949,  .949} EgoBody~\cite{SiweiZhang2021EgoBodyHB} & \textcolor{red}{\XSolidBrush} & \textcolor{green}{\Checkmark} & \textcolor{green}{\Checkmark} & \textcolor{red}{\XSolidBrush} & \textcolor{green}{\Checkmark} & \textcolor{red}{\XSolidBrush} & 15   & $<$ 50 & 20   & 153k \\
    \midrule
    SLOPER4D & \textcolor{green}{\Checkmark} & \textcolor{green}{\Checkmark} & \textcolor{green}{\Checkmark} & \textcolor{green}{\Checkmark} & \textcolor{green}{\Checkmark} & \textcolor{green}{\Checkmark} & 10   & 2k $\sim$ 13k & 12   & 100k \\
    \bottomrule
    \end{tabular}%
    
    \vspace{-2mm}
    \caption{ \textbf{Comparisons with existing datasets.}  “Global” denotes to human poses with \textbf{global translation}. The “area size” is estimated with the published data. The * indicates the data modality is only used for human self-localization rather than for human-related data.}
    \vspace{-4mm}
     \label{tab:data_compare}
 \end{table*}

\subsection{3D Human Motion Datasets}

Many datasets have been proposed with different sensors and setups to facilitate the research on 3D human pose estimation.
The H3.6M\cite{ionescu2013human3} is a large-size dataset providing synchronized video with optical-based MoCap in studio environments. To perform markerless capture in different indoor scenes, PROX~\cite{hassan2019prox} uses an RGB-D sensor to scan a single person. EgoBody~\cite{SiweiZhang2021EgoBodyHB} uses multiple RGB-D sensors to pre-scan the room and scan the interacting persons.
LiDARHuman26M~\cite{li2022lidarcap} can capture long-range human motions with static LiDAR and IMUs.
However, they are limited to static environments, human activities, and interactions. 3DPW~\cite{Marcard_2018_ECCV} is the first dataset providing 3D annotations in the wild which uses a single hand-held RGB camera to optimize human pose from IMUs for a certain period of frames. It doesn't provide accurate global translation and 3D scenes. 
HPS~\cite{guzov2021human} reconstructs the human body pose using IMUs and self-localizes it with a head-mounted camera in large 3D scenes, but it heavily relies on the pre-built map. HSC4D~\cite{Dai_2022_CVPR} removes the reliance on the pre-built map and achieves global human motion capture in large scenes. 
However, the camera in HPS and the LiDAR in HSC4D are only used to perceive the environment rather than capture human data. 
With the scene-aware dataset we proposed for global human pose estimation, we can benchmark the 3D HPE in the wild with the LiDAR or camera modalities.

\subsection{Human Localization and Scene Mapping}
Human self-localization aims at estimating the 6-DoF of the human subject in global coordinates. The image-based methods~\cite{kendall2015posenet, radwan2018vlocnet++, wang2020atloc} regress locations directly from a single image with a pre-built map. The scene-specific property makes them hard to generalize to unseen scenes. 
LiDAR is widely used in Simultaneous Localization and Mapping (SLAM)\cite{zhang2014loam, shan2018lego, bosse2012zebedee,lin2020loam} due to its robustness and low drift. To address the drift problem and improve robustness in dynamic motions, RGB cameras~\cite{zhang2015visual, shin2020dvl, seo2019tight}, IMU~\cite{shan2020lio, geneva2018lips, opromolla2016lidar}, or both~\cite{deilamsalehy2016sensor, zuo2019lic, shan2021lvi}, have been integrated with the mapping task.
Most attention has been paid to autonomous driving \cite{kitti} \cite{gskim-2020-mulran} or robotics from the third-person view and they usually do not focus on humans. To achieve self-localization, LiDAR is designed as backpacked \cite{Liu2010IndoorLA, 8736839, Karam2019DesignCA} and hand-held\cite{bauwens2016forest}. To efficiently capture human motions and reconstruct urban scenes, we utilize LiDAR with a built-in IMU (different from the IMUs for motion capture) and propose a pipeline for constructing multi-modal data.
This approach provides accurate information on human motions at both local and global levels, as well as enables mapping in large outdoor environments. 

\subsection{Global 3D Human Pose Estimation}
Most studies recover human meshes in camera coordinate~\cite{zhen2020smap, li2021hybrik} or root-relative poses\cite{kanazawaHMR18,kolotouros2019spin, kocabas2020vibe}.
Recovering global human motions in unconstrained scenes is a challenging topic in computer vision and has gained more and more research interest in recent years.
IMU sensors are widely used in commercial~\cite{roetenberg2007moven, roetenberg2009xsens} and research activities~\cite{vlasic2007practical, SIP, DIP:SIGGRAPHAsia:2018}, and are attached to body limbs to capture human motions in studio-environments.  But it suffers severe drift in the wild. 
Some methods rely on additional RGB~\cite{Marcard2016HumanPE, Marcard_2018_ECCV, dou2016fusion4d, Malleson3DV17} or pre-scan maps~\cite{guzov2021human}, or LiDAR~\cite{Dai_2022_CVPR} to complement the IMUs in large-scale scenes.
Based on human-scene interaction, some work proposed scene-aware solutions using static cameras~\cite{yi2022human, huang2022capturing} to obtain accurate and scene-natural human motions. 4DCapture\cite{MiaoLiu20204DHB} uses a dynamic head-mounted camera to self-localize and reconstruct the scene with the Struct From Motion method. However, it often fails when the illumination changes in the wild. MOVER~\cite{yi2022human} uses a single camera to optimize the 3D objects in a static scene, resulting in better 3D scene reconstruction and human motions.
GLAMR~\cite{yuan2022glamr} uses global trajectory predictions to constrain both human motions and dynamic camera poses, achieving state-of-the-art results on in-the-wild videos.
However, it lacks a benchmark for quantitatively comparing different HPE methods on a global level. To deal with this limitation, we propose \TITLE, the first large-scale urban-level human pose dataset with rich 2D/3D annotations.

\section{\TITLE~Dataset}
\label{sec:dataset_constructing}
\TITLE~collects scene-aware 4D human data with our body-worn capturing system in urban scenes. In this section, we first introduce the data acquisition in \cref{subsec:data_collecting}, second, we detail the data construction and annotation process in \cref{subsec:data_anno}, then we introduce the global optimization-based \cref{subsec:optimization} method to obtain high quality both 3D/2D data, finally, we compare our dataset in \cref{subsec:data_comapre} with the existing datasets and highlight our novelty.

\subsection{Data Acquisition}
\label{subsec:data_collecting}

\begin{figure}[!htb]
    \centering
     \includegraphics[width=0.98\linewidth]{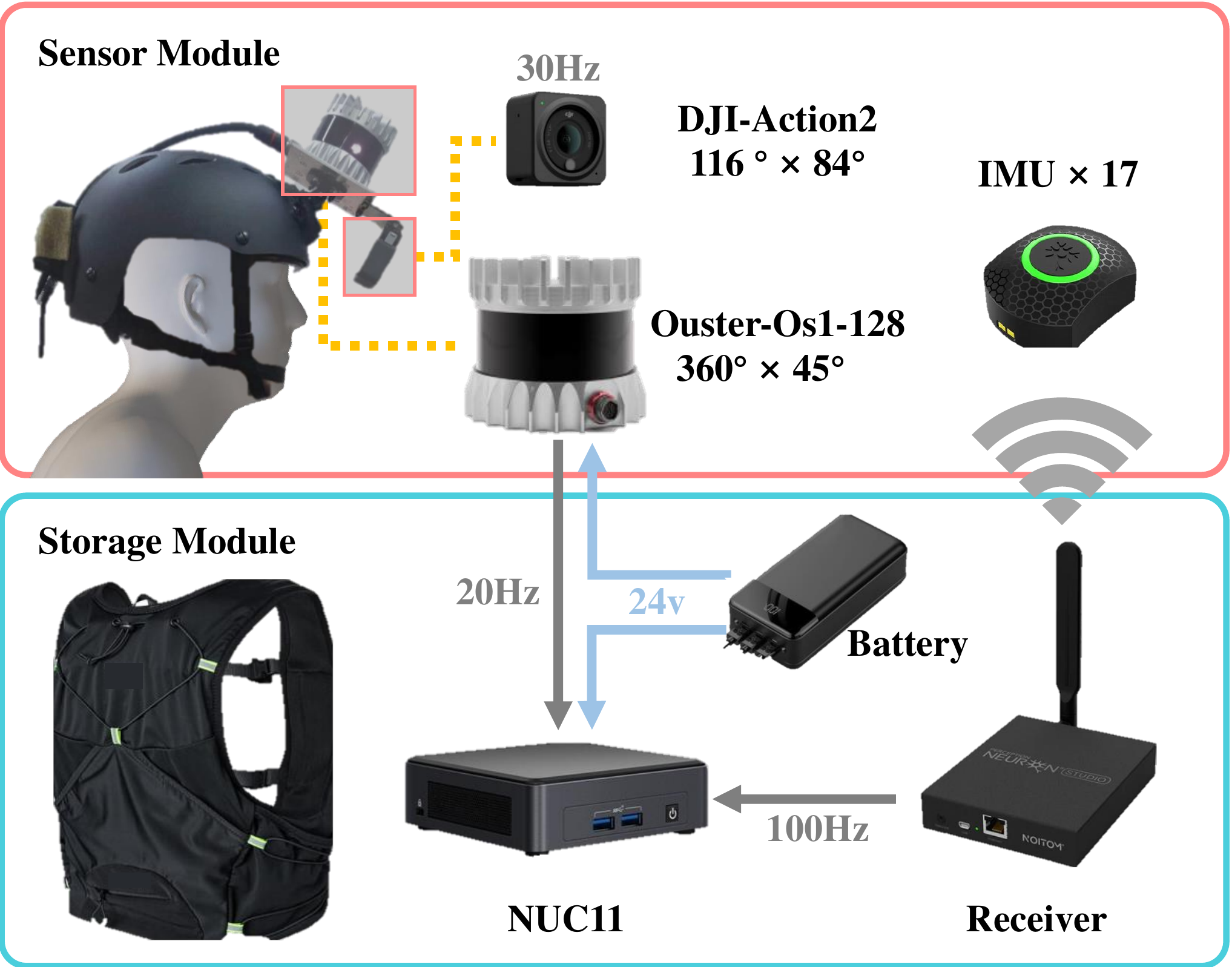}
     \vspace{-1mm}
     \caption{\textbf{Our capturing system's hardware details.} The sensor module includes a LiDAR, a camera, and 17 body-attached IMU sensors. The storage module consists of a NUC11, a receiver, and a battery in the backpack.}
     \label{fig:system_design}
    %  \vspace{-\baselineskip}
    \vspace{-5mm}
\end{figure}

\PAR{Hardware setup.}
As shown in \cref{fig:teaser}, during the data collection procedure, the scanning person follows the performer (IMUs wearer) and scans him with a LiDAR and a camera on the helmet. Additionally,  \cref{fig:system_design} shows the hardware details of our capturing system. Regarding the sensor module, the 128-beams Ouster-os1 LiDAR and the DJI-Action2 wide-range camera are rigidly installed on the helmet. To capture raw human motions, we use Noitom's inertial MoCap product, PN Studio, to attach 17 wireless IMUs to the IMU wearer's body limbs, torso, and head. 
The camera's field of view (FOV) is 116°$\times$84° and the LiDAR's FOV is 360°$\times$45°. To make the performer within the LiDAR's FOV as much as possible, we tilt the LiDAR down around 45°. 
Regarding the storage module, the scanning person's backpack places a wireless IMU data receiver, a 24V battery, and an Intel NUC11. The mini-computer NUC11 stores IMU data from the wireless receiver and point clouds from LiDAR in real time. Videos are stored locally in the camera. The LiDAR and NUC11 are both powered by the battery.

\PAR{Coordinate systems.}
Let's define three coordinate systems: 1) IMU coordinate system \{$I$\}: the origin is at LiDAR wearer's spine base at the starting time, and the $X/Y/Z$ axis is pointing left/upward/forward of the human. 2) LiDAR Coordinate system \{$L$\}: the origin is at the center of the LiDAR, and the $X/Y/Z$ axis is pointing right/forward/upward of the LiDAR. 3) Global/World coordinate system \{$W$\}: the origin is on the floor of the LiDAR wearer's starting position, and the $X/Y/Z$ axis is pointing right/forward/upward of the LiDAR wearer.

\PAR{Calibration.} 
Following the setup in \cite{wen2019toward}, we use a chessboard to calibrate the camera intrinsic $K_{in}$ and introduces a terrestrial laser scanner (TLS) to obtain accurate camera extrinsic parameter, $K_{ex}$. Due to the LiDAR point cloud being too sparse, we manually choose the corresponding points both on the 2D image and the TLS map registered to the point cloud, and then we solve the perspective-n-point (PnP) problem to obtain $K_{in}$.
For every 3D scene, the calibration $R_{WL}$, which transforms \{$L$\} to \{$W$\} is manually set to make the ground's $z$-axis upward and height to zero for the starting position.
By using singular value decomposition, the calibration $R_{WI}$, which transforms \{$I$\} to \{$W$\}, is calculated through the similarity between IMU trajectory and LiDAR trajectory on the XY plane.
\PAR{Synchronization.}
The synchronization of data from multiple sensors in human subject data is achieved through peak detection.
Before and after the capture, the subject is asked to perform jumps. Then the peak height time in IMU is automatically detected and the peak times in the LiDAR and camera data are manually identified. Finally, all modalities are aligned by the peaks and downsampled to match the LiDAR frame rate of 20 Hz.  

\subsection{Data Processing}
\label{subsec:data_anno}
\PAR{2D pose detection.} 
We use Detectron~\cite{wu2019detectron2} to detect and Deepsort~\cite{wojke2017simple} to track humans in videos. However, the tracking often fails due to the IMUs wearer entering/exiting the field of view or occlusions. To solve this problem, we manually assign the same ID for the tracked person in a video sequence. As for 3D point cloud reference, we project them on images according to the $K_{ex}$. However, due to the jitter brought by dynamic motions, the camera and the LiDAR are not perfectly rigidly connected. 
Thus, $K_{ex}$ will be further optimized in \cref{subsec:optimization}. 

\PAR{LiDAR-inertial localization and mapping.}
The LiDAR-only method often fails in mapping because of the dynamic head rotation and crowded urban environments.
Incorporating an IMU can compensate for motion distortion in a LiDAR scan $p^{L}$ and provide an accurate initial pose. 
Using a LiDAR with an integrated IMU, and by combining Kalman filter-based lidar-inertial odometry\cite{Xu2022FASTLIO2FD} with factor graph-based loop closure optimization\cite{gtsam}\cite{Kim2018ScanCE}, we successfully estimate the ego-motion of LiDAR and build the global consistency 3D scene map with n frame point clouds $P_{1:n}^L=\{p_{1}^{L},\ldots, p_{n}^{L}\}$. 
To provide accurate scene constrain in \cref{subsec:optimization}, we utilize the VDB-Fusion~\cite{vizzo2022sensors} to generate a clean scene mesh $\bm{S}$ that excludes moving objects.

\PAR{IMUs pose estimation.}
We use SMPL~\cite{smpl2015loper} to represent the human body motion ${M^I} = \varPhi (\theta^I, t^I, \beta) \in \mathbb{R}^{6890}$ in IMU coordinate space\{$I$\}, 
where pose parameter $\Theta_{1:n}^I = \{\theta_{1}^{I},\ldots, \theta_{n}^{I}\} \in \mathbb{R}^{72 \times n}$ is composed of pelvis joint's orientation $R_{1:n}^I = \{r_{1}^{I},\ldots, r_{n}^{I}\} \in \mathbb{R}^{3 \times n}$ and the other 23 joints' rotation relative to their parent joint. 
The $T_{1:n}^I = \{t_{1}^{I},\ldots, r_{n}^{I}\} \in \mathbb{R}^{3 \times n}$ is the pelvis joint's translation and $\beta \in \mathbb{R}^{10}$ is a constant value representing a person's body shape. 
$T$ and $\Theta$ are estimated by the commercial MoCap product, while $\beta$ is obtained by using IPNet~\cite{bhatnagar2020ipnet} to fit the scanned model captured by an iPhone13 Promax.
Since the IMU are accurate locally but drift globally, $T^I$ is used for raw calibration of the \{$I$\} to \{$W$\}, and the initial global motion $M = M^W = R^{WI}M^I$ will be further optimized.

\begin{figure*}[!htb]
    \centering
     \includegraphics[width=0.98\linewidth]{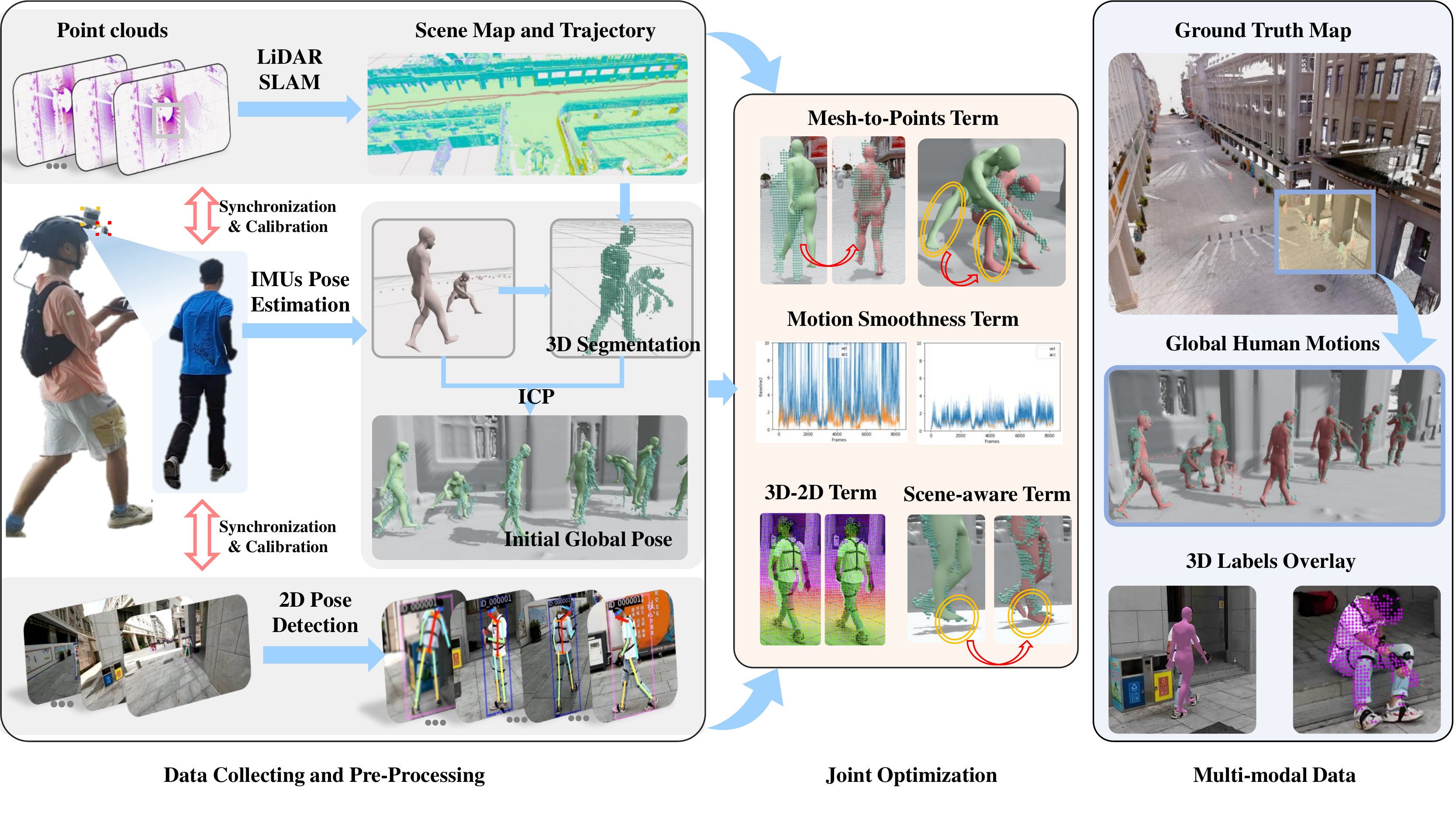}
     \vspace{-1mm}
     \caption{\textbf{The pipeline of the dataset construction.} The capturing system simultaneously collects multimodal data, including LiDAR, camera, and IMU data. Then they are further processed. A joint optimization approach with multiple loss terms is then employed to optimize motion locally and globally. As a result, we obtain rich 2D/3D annotations with accurate global motion and scene information.
     }
     \vspace{-3mm}

     \label{fig:method}
\end{figure*}

\subsection{Data Optimization}
\label{subsec:optimization}
To obtain precise and scene-plausible human motion $M$ in the world coordinate system, we use scene geometry \bm{${S}$} with several physic-based terms to perform joint optimizations to find the optimal motion $M^*$ that minimize $\mathcal{L}$. 
In a k-frame segment, the optimization is written as:
\vspace{-1mm}
\begin{equation}
	\begin{split}   
    & M^*_{1:k} = 
    \arg \min _{M_{1:k}}\mathcal{L} (M_{1:k}, \bm{S}),\\
    & \mathcal{L}_{} = 
        {\mathcal{L}}_{smt} + 
        \lambda_{sc}\mathcal{L}_{sc} + 
        {\lambda_{pri}\mathcal{L}}_{pri}+
        {\lambda_{m2p}\mathcal{L}}_{m2p}, \\
    & \mathcal{L}_{smt} = 
        \lambda_{trans} \mathcal{L}_{trans} + 
        \lambda_{orit} \mathcal{L}_{orit} + 
        \lambda_{jts} \mathcal{L}_{jts},
    \end{split}
\end{equation}

\noindent where $\mathcal{L}_{smt}$ is a smoothness term, which consists of a translation loss $\mathcal{L}_{trans}$, an orientation loss $\mathcal{L}_{orit}$, and a joints loss $\mathcal{L}_{jts}$.
$\mathcal{L}_{sc}$ is a scene-aware-contact term,
$\mathcal{L}_{pri}$ is a pose prior term, 
and $\mathcal{L}_{m2p}$ is a mesh-to-points term. 
The $\lambda_{sc}$, $\lambda_{pri}$, $\lambda_{trans}$, $\lambda_{orit}$, $\lambda_{jts}$, and $\lambda_{m2p}$ are loss terms' coefficients. $\mathcal{L}$ is minimized with a gradient descent algorithm.

\PAR{Smoothness term.} The objective of this term is to minimize the acceleration of the pelvis joint, the accelerations of the other 23 pelvis-relative joints, which is denoted as $J_{1:k} = \{\jmath_{1},\ldots, \jmath_{k}\} \in \mathbb{R}^{69 \times k}$, and the angular velocity of all joints to smooth human movements. 

\PAR{Scene-aware contact term.} we compare the movement of every foot vertices in IMU motions ${M}_k^I$ and label the foot as stable if its velocity is less than 0.1 $m/s^2$. Finally, the Chamfer Distance (CD) between this foot and its closest surface is expressed as the scene contact loss $\mathcal{L}_{sc}$.

\PAR{Pose prior term.} The poses estimated by IMUs are roughly accurate but will likely cause some misalignments to the end of the body limb due to the accumulating error.
Hence, $\mathcal{L}_{prior}$ is used to constrain the $\Theta$ close to the initial value at the beginning of the optimization. 

\PAR{Mesh-to-points term.}
The point cloud $p^L$ from the moving LiDAR provides strong prior depth information. However, though the SMPL mesh is watertight and complete, the human points are sparse and partial, which makes the registration methods such as ICP, not ideal as expected.
To address this issue, we propose a viewpoint-based mesh-to-point loss function $\mathcal{L}_{m2p}$. First, we remove the hidden SMPL mesh faces from the LiDAR's viewpoint. Then we sample points, denoted as $P'\,\!_{1:k} = \{p'\,\!_{1},\ldots, p'\,\!_{k}\}$, from the remaining faces by LiDAR resolution. The loss is defined as the Chamfer Distance from $P'\,\!_{1:k}$ to $P_{1:k}$. 

All loss terms functions are detailed as follows:

\vspace{-4mm}
\begin{equation}
	\begin{split} 
        & \mathcal{L}_{trans} = 
        \frac{1}{k-2}\sum_{i=1}^{k-2}\|t_{i+2} - 2t_{i+1} + t_{i}\|_2^2,\\ 
        & \mathcal{L}_{jts} = 
        \frac{1}{k-2} \sum_{i=1}^{k-2}\|{\jmath}_{i+2} - 2{\jmath}_{i+1} + {\jmath}_{i}\|_2^2,\\ 
        & \mathcal{L}_{\text {orit}} = 
        \frac{1}{k-1}\sum_{i=1}^{k-1} 
        \|r_{i+1} - r_{i}\|_{2}^2, \\
        & \mathcal{L}_{\text {pri}} = 
        \frac{1}{k}\sum_{i=1}^{k} 
        \|\theta_{i} - R^{WI}\theta_{i}^{I}\|_{2}^2, \\
        & \mathcal{L}_{m2p} = 
        \frac{1}{k}\sum_{j=1}^{k} 
        (\frac{1}{|p'\,\!_{i}|}\sum_{\hat{p'\,\!}\in p'\,\!_{i}}\min_{\hat{p} \in p_{i}}\|\hat{p} - \hat{p'\,\!}\|_{2}^{2}).
    \end{split}
    \vspace{-4mm}
\end{equation}

\PAR{Camera extrinsic optimization.}
We aim to optimize extrinsic parameters $K_{ex}$ for every frame by minimizing the $\mathcal{L}_{cam}$, which comprises of the keypoints loss $\mathcal{L}_{kpt}$ and the bounding box loss $\mathcal{L}_{box}$. 
The $\mathcal{L}_{kpt}$ measures the mean square error (MSE) between the 2D human keypoints $kpt^{2d}$ in the image and the 3D human keypoints $kpt^{3d}$ of the optimized SMPL model projected to the image with $K_{ex}$; 
the $\mathcal{L}_{box}$ computes the Intersection over Union(IoU) loss between the 2D human bounding box $box^{2d}$ in the image and the 3D human bounding box $box^{3d}$ projected to the image with $K_{ex}$. 

\vspace{-4mm}
\begin{equation}
	\begin{split}   
    K_{ex}^* = &
    \arg \min _{K_{ex} \in SE(3)} \mathcal{L}_{cam}(K_{ex}), \\
    \mathcal{L}_{cam} = &
    \lambda_{kpt} \mathcal{L}_{kpt}(kpt^{3d}, kpt^{2d}, K_{ex}) + \\
    & \lambda_{box} \mathcal{L}_{box}(box^{3d}, box^{2d}, K_{ex}), \\
    \end{split}
    \vspace{-2mm}
\end{equation}
\noindent where $\lambda_{kpt}$ and $\lambda_{box}$ are constant coefficients.

\begin{figure*}[!htb]
    \centering

    \includegraphics[width=0.99\linewidth]{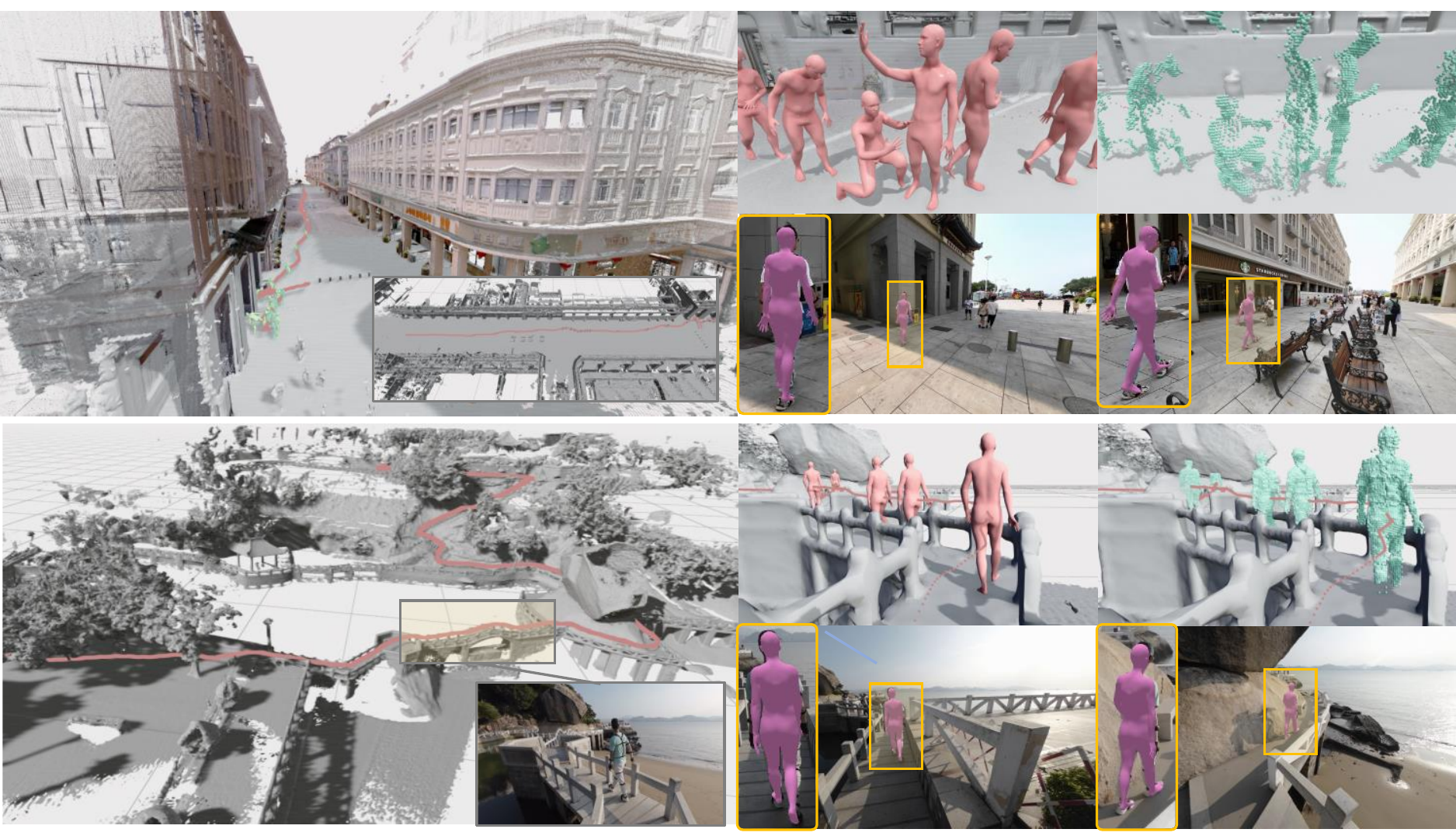}  
    \vspace{-1mm}
    \caption{The diverse scenes and activities of our dataset. The images in the left column are our reconstructed scenes with human trajectories overlaid on them. The right images are the SMPL meshes overlaid on images / point clouds / scenes.
    }

    \vspace{-3mm}
    \label{fig:dataset}
 \end{figure*}

\subsection{Dataset Comparison}
\label{subsec:data_comapre}

\TITLE~is the first large-scale urban-level human pose dataset with multi-modal capture data and rich human-scene annotations for GHPE.
The head-mounted LiDAR and camera are utilized to simultaneously record the IMU-wearer's activities, including running outside, playing football, visiting, reading, climbing/descending stairs, discussing, borrowing a book, greeting, etc.

The dataset consists of \numberseq sequences from \numberperson human subjects in \numberscene locations. 
There are a total of \framelidar LiDAR frames, \framevideo video frames, and \framemocap IMU-based motion frames captured over a total distance of more than 8 $km$ and an area of up to 13,000 $m^2$. The results of our dataset are shown in \cref{fig:dataset}.
For the captured person, we provide the segmentation of 3D points from LiDAR frames and 2D bounding boxes from images synchronized with LiDAR. We also provide 3D pose annotations with SMPL format. Compared to other datasets~\cref{tab:data_compare}, it is worth mentioning that \TITLE~ provides the 3D scene reconstructions and accurate global translation annotations, allowing us to quantitatively study the scene-aware global pose estimation from both LiDAR and monocular videos. In addition to the dense 3D point cloud map reconstructed from the LiDAR, \TITLE~ provides the high-precision colorful point cloud map from a Terrestrial Laser Scanner (Trimble TX5) for better visualization and map comparison.

\section{Experiments}
\label{sec:experiments}

\begin{figure}[!htb]
    \centering
     \includegraphics[width=0.98\linewidth]{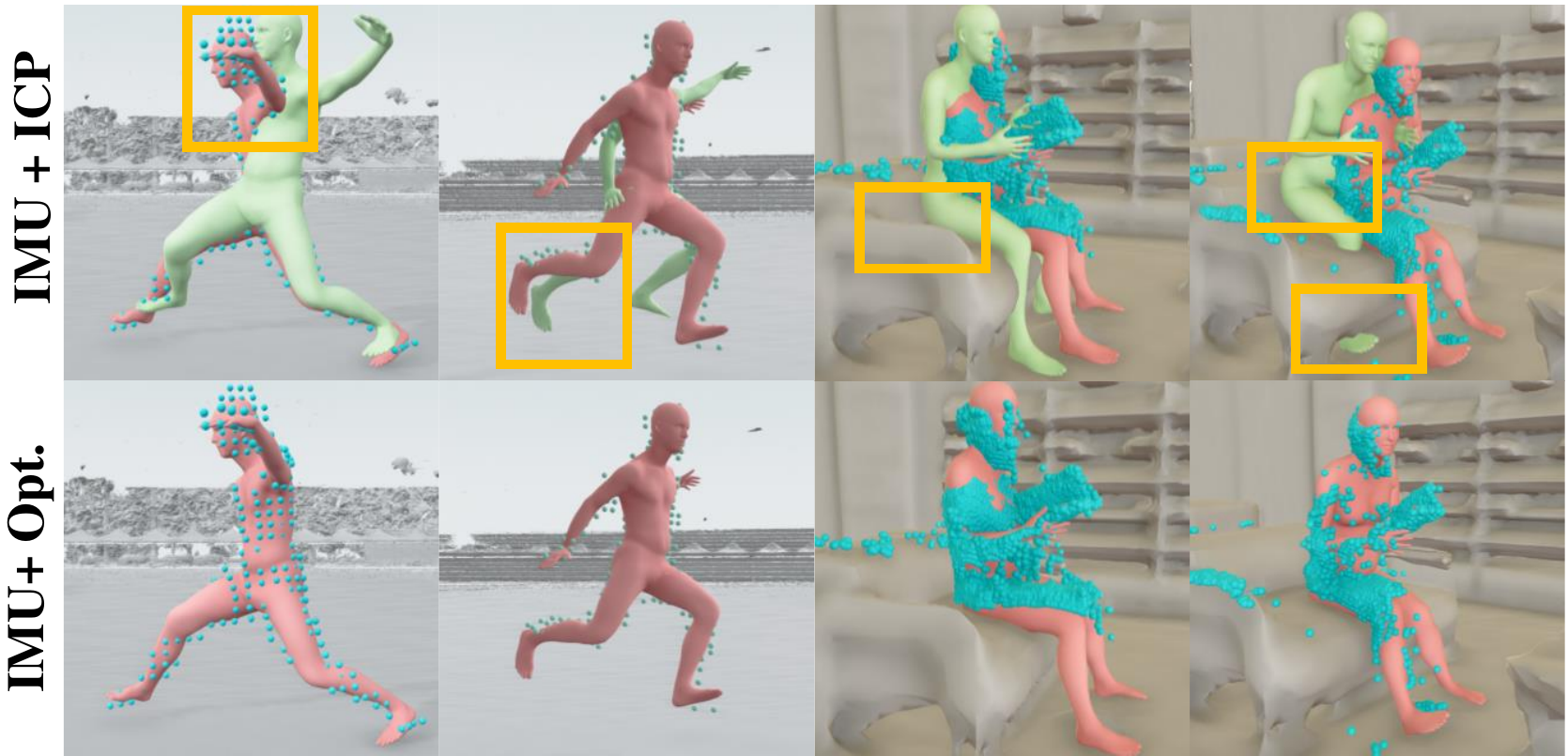}
     \vspace{-1mm}
     \caption{
     Comparison between our optimization results (\textbf{\textcolor[rgb]{0.937,0.686,0.69}{red SMPL}}) and the ICP results (\textcolor[RGB]{141,202,108}{\textbf{green SMPL)}}. It shows the red SMPL aligns better with the \textcolor[rgb]{0.133, 0.792, 0.796}{\textbf{cyan human points}} than the green SMPL.}
     \label{fig:eval_opt}

    \vspace{-3mm}
\end{figure}

\begin{figure}[!htb]
    \centering
     \includegraphics[width=0.98\linewidth]{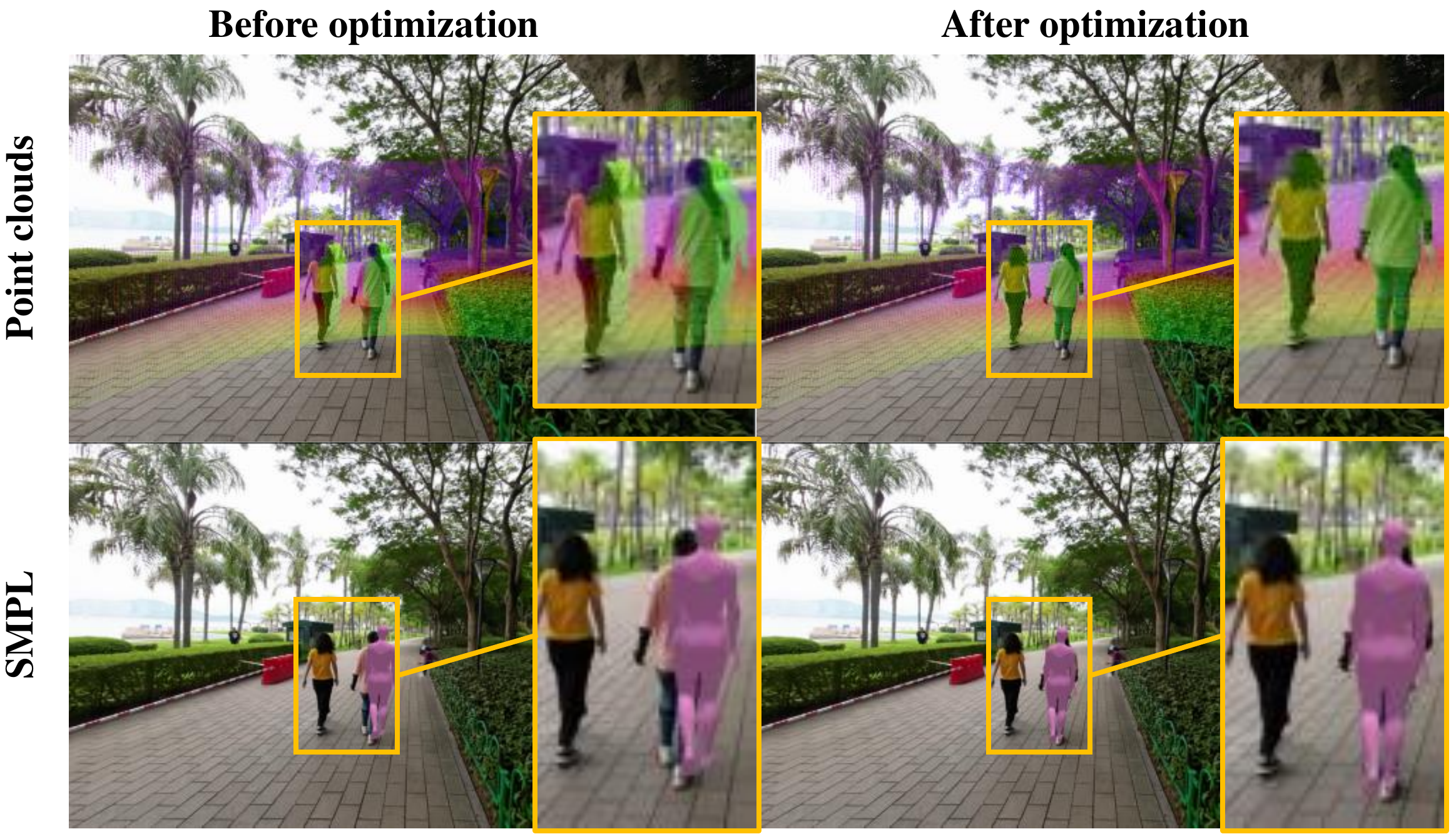}
     \vspace{-3mm}
     \caption{The comparison before (\textbf{left}) and after (\textbf{right}) extrinsic optimization by projecting the point clouds (upper) and SMPL (lower) are projected onto the image.}
     \label{fig:eval_cali}

    \vspace{-4mm}
\end{figure}

 In this section, we first evaluate \TITLE~Dataset qualitatively, indicating that our dataset is solid enough to benchmark new tasks. Then we perform a cross-dataset evaluation to further assess our dataset's novelty on two tasks: LiDAR-based 3D HPE and camera-based 3D HPE. Finally, we introduce the new benchmark, GHPE, and perform experiments on GLAMR. More quantitative evaluations and experiments are in the supplementary material.

\PAR{Training/Test splits.} We split our data into training and test sets for LiDAR/Camera-based pose estimation. The training set of \TITLE~ contains eleven sequences of data with a total of 80k LiDAR frames and corresponding RGB frames. The test set has four sequences of data with around 20k LiDAR frames and corresponding RGB frames.

For global pose estimation, we select three challenging scenarios for evaluation. The first one is a single-person football training scenario with highly dynamic motions. The second one is running along a coastal runway. The third one is a garden tour involving daily motions.

\PAR{Evaluation metrics.}
For 3D HPE, we employ Mean per joint position error (MPJPE) and Procrustes-aligned MPJPE (PA-MPJPE) for evaluation. MPJPE is the mean euclidean distance between the ground-truth and predicted joints. PA-MPJPE first aligns the predicted joints to the ground-truth joints by carrying out rigid transformation based on Procrustes analysis and then calculates MPJPE. 
For global trajectory evaluation, we utilize Absolute Trajectory Error (ATE) and the Relative Pose (the pose refers to orientation here) Error (RPE) in visual SLAM systems \cite{grupp2017evo}, where the ATE is well-suited for measuring the global localization and, in contrast, the RPE is suitable for measuring the system's drift, for example, the drift per second. 
Global MPJPE (G-MPJPE) is MPJPE calculated by placing the SMPL model in the global coordinates.

\PAR{Qualitative evaluation.} For the human pose qualitative evaluation, we project the SMPL to the image and visualize the 3D human with corresponding LiDAR points in 3D space (shown in \cref{fig:dataset}). The results demonstrate that the 3D human mesh aligns well with 3D environments and 2D images. As a large-scale urban-level human pose dataset, \TITLE~ provides multi-modal capture data and rich human-scene annotations, as well as diverse challenging human activities in large scenes.
To evaluate our optimization method, we first compare our method with the results from ICP. As shown in \cref{fig:eval_opt}, the scene-aware constraints and human mesh-to-points constraint efficiently optimize the local poses, global translation, and even the orientation error from IMU.
To show the effectiveness of the camera extrinsic optimization,  we report the results in \cref{fig:eval_cali}. The 2D projecting error was visually lowered after optimization.

\subsection{Cross-Dataset Evaluation}

\begin{table}[!htb]
    \centering
    \footnotesize

\begin{tabular}{l|rr|rr}
    \hline
    \multirow{2}{*}{\begin{tabular}[c]{@{}l@{}}\diagbox{Train\quad}{Test}\end{tabular}} & \multicolumn{2}{c|}{LH26M } & \multicolumn{2}{c}{Ours} \\ 

        & MPJPE  & PA-MPJPE & MPJPE  & PA-MPJPE \\ 
        \hline
    LH26M      & \textbf{79.3} & \textbf{67.0} & 228.7 & 149.9 \\
    Ours       & 212.3 & 128.3  & 86.1  & 65.1 \\
    LH26M + Ours & 85.5  & 72.0   & \textbf{79.2}  & \textbf{60.1} \\
    \hline

    \multicolumn{5}{c}{(a) LiDAR-based 3D pose estimation with LiDARCap~\cite{li2022lidarcap}.} \\
    \multicolumn{5}{c}{} \\

    \hline
    \multirow{2}{*}{\begin{tabular}[c]{@{}l@{}}\diagbox{Train\quad}{Test}\end{tabular}} & \multicolumn{2}{c|}{3DPW} & \multicolumn{2}{c}{Ours} \\ 

        & MPJPE & PA-MPJPE & MPJPE  & PA-MPJPE \\ \hline
    VIBE~\cite{kocabas2020vibe} & 93.5  & 56.5 & 102.5 & 66.2 \\
    Ours + AMASS & 124.3 & 66.8 & \textbf{86.6}  & \textbf{52.4} \\
     w. 3DPW     & \textbf{83.0} & \textbf{52.0} & 90.0  & 58.3 \\

    \hline
    HbryIK~\cite{li2021hybrik}  & 88.7  & 49.3  & 104.9 & 57.0          \\
    w. Ours   & 87.3 & 49.2 & 67.6 & 44.2    \\
    w. 3DPW   & \textbf{71.3} & \textbf{41.8} & 75.8 & 50.0 \\
    w. 3DPW + Ours & 76.4 & 46.7 & \textbf{66.2} & \textbf{42.8}  \\

    \hline
    \multicolumn{5}{c}{(b)  Camera-based 3D pose estimation.} \\
    \end{tabular}%
    \vspace{-1mm}
    \caption{Cross-dataset evaluation results with different modalities. The LH26M in (a) refers to LiDARHuman26M dataset from LiDARcap. VIBE is pre-trained on AMASS~\cite{mahmood2019amass}, MPI-INF-3DHP~\cite{mono-3dhp2017} InstaVariety~\cite{humanMotionKZFM19}, PoseTrack~\cite{andriluka2018posetrack}, PennAction~\cite{zhang2013actemes}. HbryIK is pre-trained on H36M, MPI-INF-3DHP, MSCOCO~\cite{lin2014microsoft}}
    \label{tab:cross}
    \vspace{-3mm}
 \end{table}

We evaluate root-relative 3D human pose estimation with different modalities, namely the LiDAR and the camera. 3DPW is an in-the-wild human motion dataset that is most related to us. With VIBE, we cross-evaluated our dataset's camera modal by using 3DPW.
LiDARHuman26M is a lidar-based dataset for long-range human pose estimation. We can cross-evaluate our dataset's LiDAR modal with it.
\cref{tab:cross}(a) shows the evaluation results on LiDAR-based 3D pose estimation task and \cref{tab:cross}(b) shows the results on camera-based 3D pose estimation. Taking the results from \cref{tab:cross}(a), for example, when the model is trained from another dataset only,  the errors are the largest.
But the error will be further reduced by around 60\% when training on LiDARHuman26M and our dataset together. 
It suggests a domain gap exists between different LiDAR sensors, and both datasets complement each other.
The results of another task show that the pre-trained VIBE model generalizes better on 3DPW than on our dataset. But the error on 3DPW increases when finetuned on our dataset, while the error decreases on our dataset. This suggests that the pre-trained model complements \TITLE~better than the opposite. Comparing the results across different modalities, the error on our dataset from the method trained on mixed LiDAR point cloud datasets is 13\% lower than the method trained on the images.

\subsection{Benchmark on Global Human Pose Estimation}

\begin{table}[htb]
    \centering
    \footnotesize
    \vspace{-3mm}
\begin{tabular}{llrrrr}
    \toprule
    Scene & Metric & \multicolumn{1}{l}{RMSE $\downarrow$} & \multicolumn{1}{l}{$mean$} & \multicolumn{1}{l}{$std.$} & \multicolumn{1}{l}{$max$} \\
    \midrule
    Football & ATE & 3.26 & 2.85 & 1.58 & 11.83 \\
    Running001 & ATE & 29.48 & 25.55 & 14.72 & 56.07 \\
    Garden001 & ATE & \textbf{2.86} & 2.57 & 1.26 & 6.55 \\
    \midrule
    Football & RPE & 0.08 & 0.06 & 0.05 & 1.34 \\
    Running001 & RPE & 0.40 & 0.35 & 0.19 & 1.04 \\
    Garden001 & RPE & \textbf{0.06} & 0.04 & 0.04 & 0.71 \\
    \bottomrule
    \end{tabular}%
    \vspace{-2mm}

    \caption{Global trajectory evaluation of GLAMR. Unit: $m$.}
    \vspace{-3mm}

    \label{tab:ghpe}
\end{table}

\begin{table}[htb]
    \centering
    \footnotesize
    \vspace{-3mm}
\begin{tabular}{lrrrr}
    \toprule
    Scene & \multicolumn{1}{l}{Scale} & \multicolumn{1}{l}{MPJPE $\downarrow$} & \multicolumn{1}{l}{PA-MPJPE $\downarrow$} & \multicolumn{1}{l}{G-MPJPE $\downarrow$} \\
    \midrule
    Football & 11.83 & 264.6 & 118.5 & 5268.7 \\
    Running001 & 56.07 & 652.1 & 119.6 & 32329.3 \\
    Garden001 & 6.55 & \textbf{139.4} & \textbf{86.3} & \textbf{4407.0} \\
    \bottomrule
    \end{tabular}%
    \vspace{-2mm}
    
    \caption{GHPE results from GLAMR. Unit: $mm$.}
    \vspace{-3mm}
    \label{tab:ghpe2}
\end{table}

\begin{figure}[!htb]
    \centering
     \includegraphics[width=0.98\linewidth]{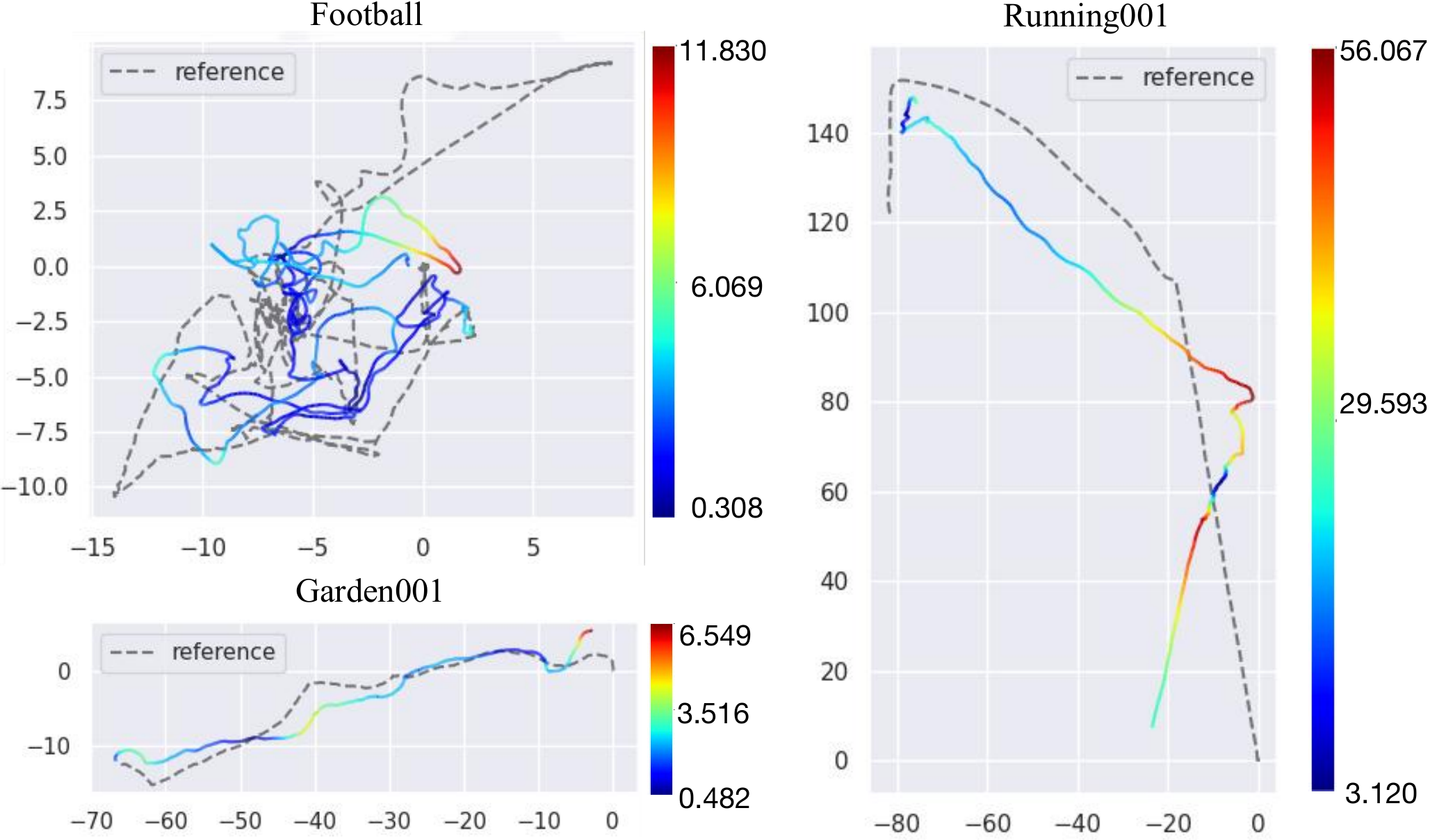}
     \vspace{-2mm}
     \caption{The ATE error mapped on the GT trajectory. The color represents the error according to the color bar.}
     \label{fig:glamr}

    \vspace{-4mm}
\end{figure}

In this subsection, we benchmark the GHPE task of GLAMR~\cite{yuan2022glamr} on \TITLE. GLAMR is a global occlusion-aware method for 3D global human mesh recovery from dynamic monocular cameras. For the scale uncertainty of the monocular camera, we compute the affine matrix from the estimated trajectory to the ground truth trajectory and rotate, translate and scale the estimated trajectory before error computation.

\cref{tab:ghpe} reports the global trajectory error with ATE and RPE, \cref{tab:ghpe2} reports the global human pose metric, and \cref{fig:glamr} shows the ATE error mapped on GT trajectory.
Comparing the results on the three scenes, the \textit{football} and \textit{Garden001} have a significantly lower RPE in the global scene.
In comparison, GLAMR performs the worst on the running scene, with an ATE's RMSE of 29.48 m. This scene has the largest area size and the highest human pace. GLAMR achieves a low PA-MPJPE of 86.3mm on \textit{Garden001}, a sequence with daily walking and visiting motions. It's the first time that we have tested the GPHE on such large outdoor scenes. GLAMR achieves relatively better results on daily human motion while performing worse on high-dynamic activities in the wild. The interesting point is that the trajectory tendency is pretty similar to the reference, even in dynamic football training motions, which demonstrates the ability of GLAMR to be a baseline. It is expected that more research will focus on GHPE in real-world interactive scenarios, and the experiments show our \TITLE's potential to promote urban-level GHPE research.

\section{Discussions}
\label{sec:discussion}

\PAR{Limitations.} 
Firstly, SLOPER4D is limited to single-person capture though it perceives multiple-person data. Secondly, the camera and LiDAR are not synchronized online, causing tedious offline work if the camera loses frames even with a low time offset (\textless50 $ms$). Finally, texture information from the camera is not fully exploited for color and texture reconstruction of scenes and humans.
In our future work, we will propose an online synchronization algorithm and extend our work to multiple-person capturing.

\PAR{Conclusions.} 
We propose the first large-scale urban-level human pose dataset with multi-modal capture data and rich human-scene annotations. Based on our proposed new dataset, we benchmark two critical tasks, camera-based 3D HPE and LiDAR-based 3D HPE. SLOPER4D also benchmarks the GHPE task. The results demonstrate the potential of SLOPER4D in boosting the development of these areas. 

Our work contributes to extending motion capture to large global scenes based on the current methods and datasets. We hope this work will foster future creation and interaction in urban environments. 

{Acknowledgements.} We thank Zhiyong Wang for helping us incorporate FAST-LIO2 into our mapping system.
This work was supported in part by the National Natural Science Foundation of China (No.62171393, No.62206173), 
the Fundamental Research Funds for the Central Universities (No.20720220064), 
the open fund of PDL (WDZC20215250113, 2022-KJWPDL-12),
and FuXiaQuan National Independent Innovation Demonstration Zone Collaborative Innovation Platform (No.3502ZCQXT2021003).
 We also acknowledge support from Shanghai Frontiers Science Center of Human-centered Artificial Intelligence (ShangHAI). 

{\small
\bibliographystyle{ieee_fullname}
\bibliography{egbib}
}

\end{document}